%% file: main.tex
\pdfoutput=1
\documentclass[11pt]{article}

\usepackage{acl}
\usepackage{times}
\usepackage{latexsym}
\usepackage[T1]{fontenc}
\usepackage[utf8]{inputenc}
\usepackage{microtype}

\usepackage{amsfonts}
\usepackage{amsmath}
\usepackage{amssymb}
\usepackage{mathtools}
\usepackage{subcaption}
\usepackage{xspace}
\usepackage{adjustbox}

\usepackage{enumitem}
\setlist{nolistsep} 
\setlist[itemize]{leftmargin=*}

\def \MODELNAME {AdaLoGN}
\def \MODELNAMEFULL {Adaptive Logic Graph Network}

\newcommand{\sigmoid}{\ensuremath{\mathtt{sigmoid}}\xspace}
\newcommand{\relu}{\ensuremath{\mathtt{ReLU}}\xspace}
\newcommand{\leakyrelu}{\ensuremath{\mathtt{LeakyReLU}}\xspace}
\newcommand{\resbigru}{\ensuremath{\mathtt{Res\text{-}BiGRU}}\xspace}
\newcommand{\linear}{\ensuremath{\mathtt{linear}}\xspace}
\newcommand{\softmax}{\ensuremath{\mathtt{softmax}}\xspace}
\newcommand{\idx}{\ensuremath{\mathtt{idx}}\xspace}

\title{AdaLoGN: Adaptive Logic Graph Network for\\Reasoning-Based Machine Reading Comprehension}

\author{Xiao Li \and Gong Cheng \and Ziheng Chen \and Yawei Sun \and Yuzhong Qu \\
         State Key Laboratory for Novel Software Technology, Nanjing University, Nanjing, China \\
         \texttt{\{xiaoli.nju,chenziheng,ywsun\}@smail.nju.edu.cn} \\ \texttt{\{gcheng,yzqu\}@nju.edu.cn}}

\begin{document}
\maketitle
\begin{abstract}
Recent machine reading comprehension datasets such as ReClor and LogiQA require performing logical reasoning over text. Conventional neural models are insufficient for logical reasoning, while symbolic reasoners cannot directly apply to text. To meet the challenge, we present a neural-symbolic approach which, to predict an answer, passes messages over a graph representing logical relations between text units. It incorporates an adaptive logic graph network (AdaLoGN) which adaptively infers logical relations to extend the graph and, essentially, realizes mutual and iterative reinforcement between neural and symbolic reasoning. We also implement a novel subgraph-to-node message passing mechanism to enhance context-option interaction for answering multiple-choice questions. Our approach shows promising results on ReClor and LogiQA.
\end{abstract}

\input{1-introduction}
\input{3-approaches}
\input{4-experiments}
\input{4.5-related_work}
\input{5-conclusion}

\section*{Acknowledgements}
This work was supported in part by the NSFC (62072224) and in part by the Beijing Academy of Artificial Intelligence (BAAI).

\bibliographystyle{acl_natbib}
\bibliography{main}

\end{document}

%% file: 1-introduction.tex
\section{Introduction}
\label{sec:intro}

Machine reading comprehension (MRC) has drawn much research attention. Early MRC datasets are not difficult for state-of-the-art neural methods. Indeed, BERT~\cite{bert} has outperformed humans on SQuAD~\cite{squad}. Recent datasets become more challenging. For example, ReClor~\cite{reclor} and LogiQA~\cite{logiqa} require understanding and reasoning over logical relations described in text, where neural methods showed unsatisfactory performance.

\begin{figure}[t]
    \centering
    \includegraphics[width=\linewidth]{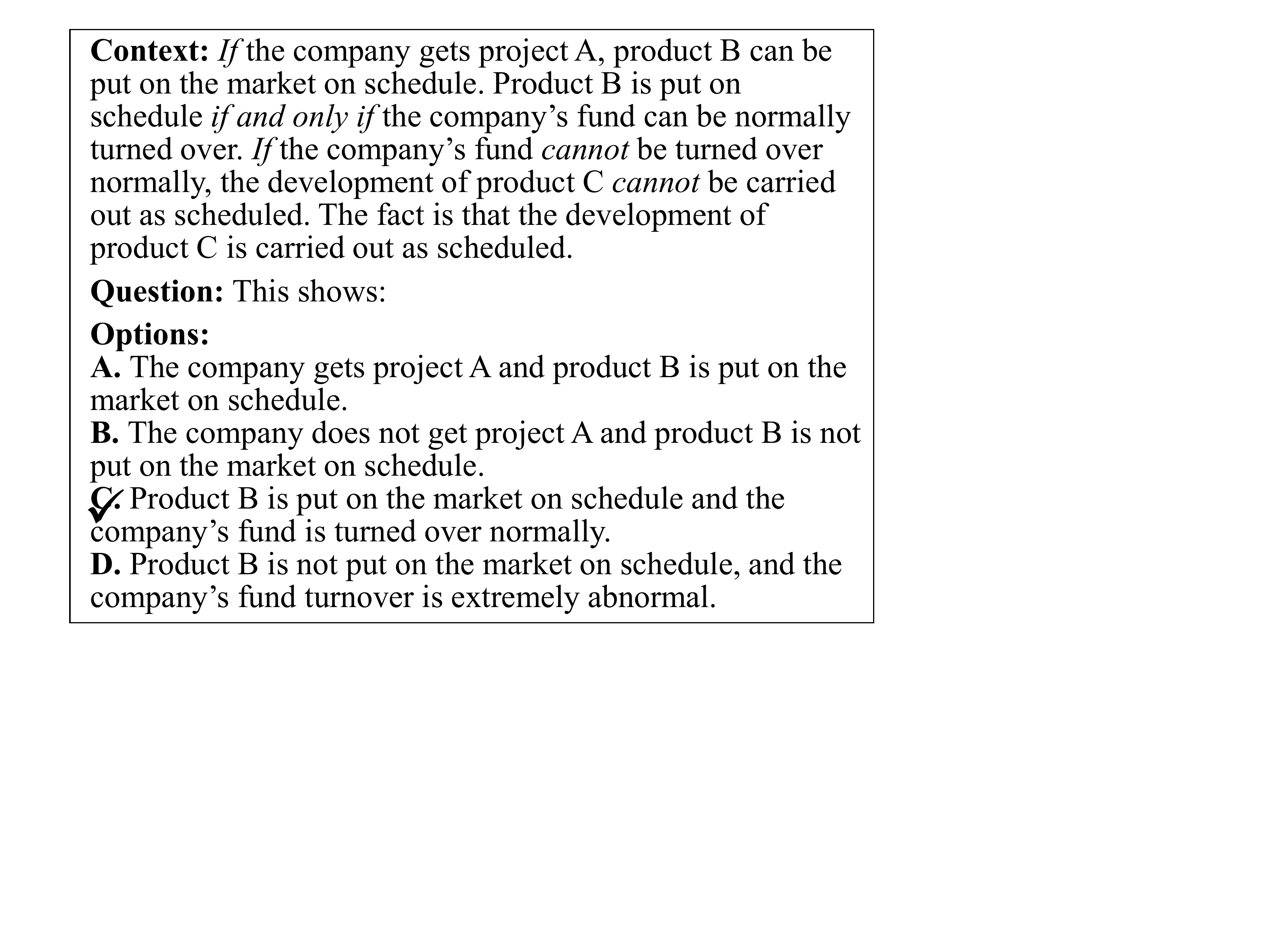}
    \caption{An example MRC task (adapted from a task in LogiQA). Logical connectives are highlighted in italics. \checkmark marks the correct answer.}
    \label{fig:example}
\end{figure}

For instance, consider the MRC task in Figure~\ref{fig:example}. The context consists of a set of textual propositions describing logical relations between elementary discourse units (EDUs)~\cite{RST}. For example, the first sentence describes an implication between two EDUs: ``the company gets project~A'' implies that ``product~B can be put on the market on schedule''. With the help of propositional calculus, humans can formalize propositions and then apply inference rules in propositional logic to prove the proposition in option~C. However, how can machines solve such a task?

\paragraph{Existing Methods and Limitations}
To solve it, conventional neural models are insufficient for providing the required reasoning capabilities, while symbolic reasoners cannot directly apply to unstructured text. One promising direction is to consider a neural-symbolic solution, such as the recent DAGN method~\cite{dagn}. It breaks down the context and each option into a set of EDUs and connects them with discourse relations as a graph. Then it performs graph neural network (GNN) based reasoning to predict an answer.

However, we identify two limitations in this method. \textbf{L1:} Despite the graph representation, it is predominantly a neural method over discourse relations. It is debatable whether the required symbolic reasoning over logical relations (e.g.,~implication, negation) can be properly approximated. \textbf{L2:} The graph is often loosely connected and composed of long paths. Node-to-node message passing implemented in existing GNN models~\cite{gcn,rgcn,gat} is prone to provide insufficient interaction between the context and the option, which is critical to answering a multiple-choice question.

\paragraph{Our Approach.}
While we follow the general framework of DAGN, i.e.,~graph construction and then graph-based reasoning, we overcome its two limitations with a novel neural-symbolic approach.

\begin{figure}[t]
     \centering
     \begin{subfigure}{\linewidth}
         \centering
         \includegraphics[width=0.9\linewidth]{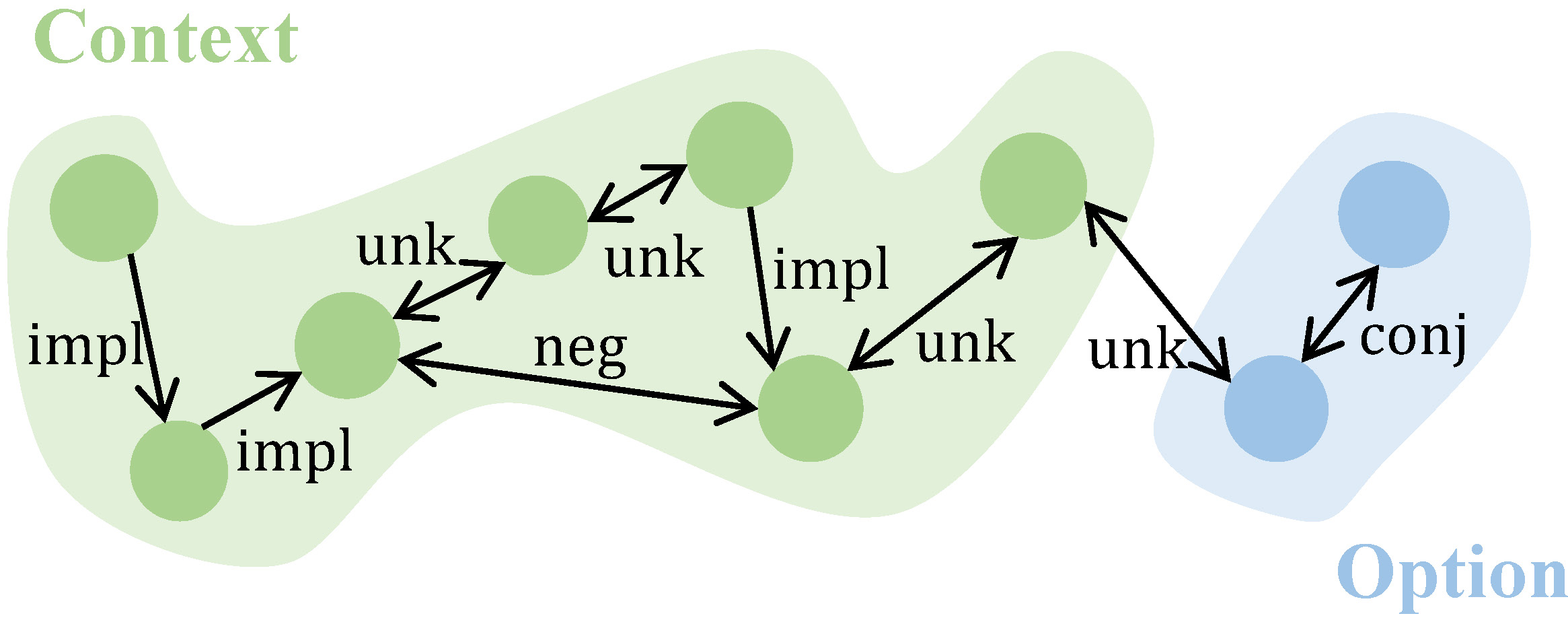}
         \caption{Raw TLG.}
         \label{fig:graph_example_a}
     \end{subfigure}
     \\
     \begin{subfigure}{\linewidth}
         \centering
         \includegraphics[width=0.9\textwidth]{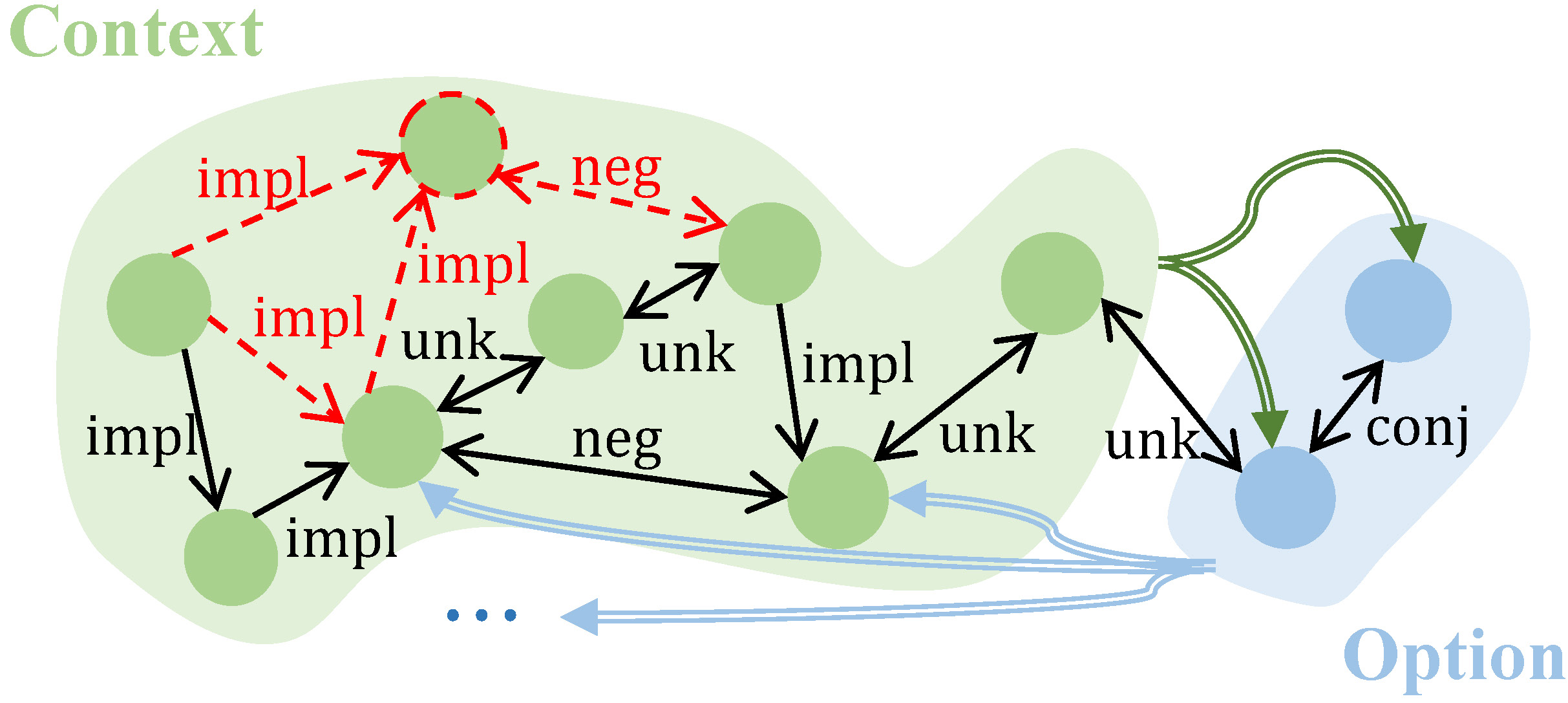}
         \caption{Extended TLG. Dashed nodes and edges represent adaptively inferred EDUs and logical relations, respectively. Double edges represent subgraph-to-node message passing.}
         \label{fig:graph_example_b}
     \end{subfigure}
     \caption{Two TLGs for exemplifying our approach. For readability, we omit \texttt{rev} edges.}
     \label{fig:graph_example}
\end{figure}

\begin{figure}[t]
    \centering
    \includegraphics[width=\linewidth]{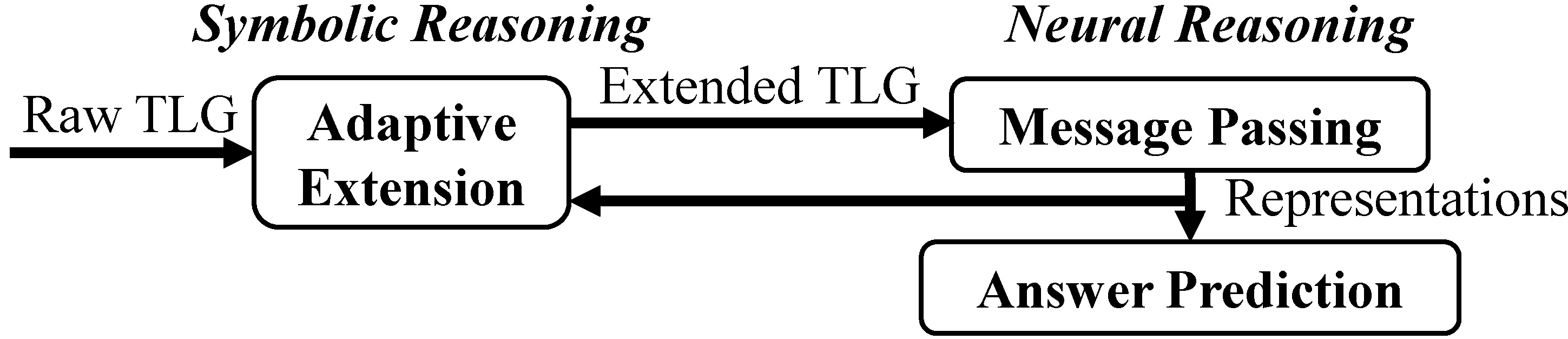}
    \caption{Our main idea: mutual and iterative reinforcement between symbolic and neural reasoning.}
    \label{fig:mini_architecture}
\end{figure}

To address~L1, Figure~\ref{fig:mini_architecture} sketches out our idea. Specifically, we propose to construct a text logic graph (TLG) representing EDUs and their logical relations as opposed to discourse relations, so we can \emph{explicitly perform symbolic reasoning} to extend the TLG with inferred logical relations, as illustrated in Figure~\ref{fig:graph_example}. The inferred relations may provide crucial connections to be used in the subsequent graph-based message passing, i.e.,~\emph{symbolic reasoning reinforces neural reasoning}.

Further, while trivially computing and admitting the deductive closure may extend the TLG with irrelevant connections which would mislead message passing, we leverage signals from neural reasoning to adaptively admit relevant extensions, i.e.,~\emph{neural reasoning reinforces symbolic reasoning}.

Moreover, we \emph{iterate the above mutual reinforcement} by restarting inference in each iteration with signals from the previous iteration to accommodate corrections to the reasoning process and allow sufficient neural-symbolic interaction.

To address~L2, we aggregate the information in the context subgraph of TLG and employ a novel subgraph-to-node message passing mechanism to \emph{enhance the interaction from the holistic context subgraph to each node in the option subgraph, and vice versa}, as illustrated in Figure~\ref{fig:graph_example_b}.

We incorporate the above two ideas into our new \MODELNAMEFULL\ (\MODELNAME). To summarize, our technical contributions include
\begin{itemize}
    \item a novel neural-symbolic approach where neural and symbolic reasoning mutually and iteratively reinforce each other, and
    \item a novel aggregation-based enhancement of message passing in graph-based neural reasoning.
\end{itemize}

\paragraph{Outline.}
We elaborate our approach in Section~\ref{sec:approach}, present experiments in Section~\ref{sec:experiments}, discuss related work in Section~\ref{sec:related_work}, and conclude in Section~\ref{sec:conclusion}.

Our code is available on GitHub: \url{https://github.com/nju-websoft/AdaLoGN}.

%% file: 3-approaches.tex
\section{Approach}
\label{sec:approach}

A MRC task $\langle c,q,O \rangle$ consists of a context~$c$, a question~$q$, and a set of options~$O$. Only one option in~$O$ is the correct answer to~$q$ given~$c$. The goal of the task is to find this option.

\begin{figure}[t]
    \centering
    \includegraphics[width=\linewidth]{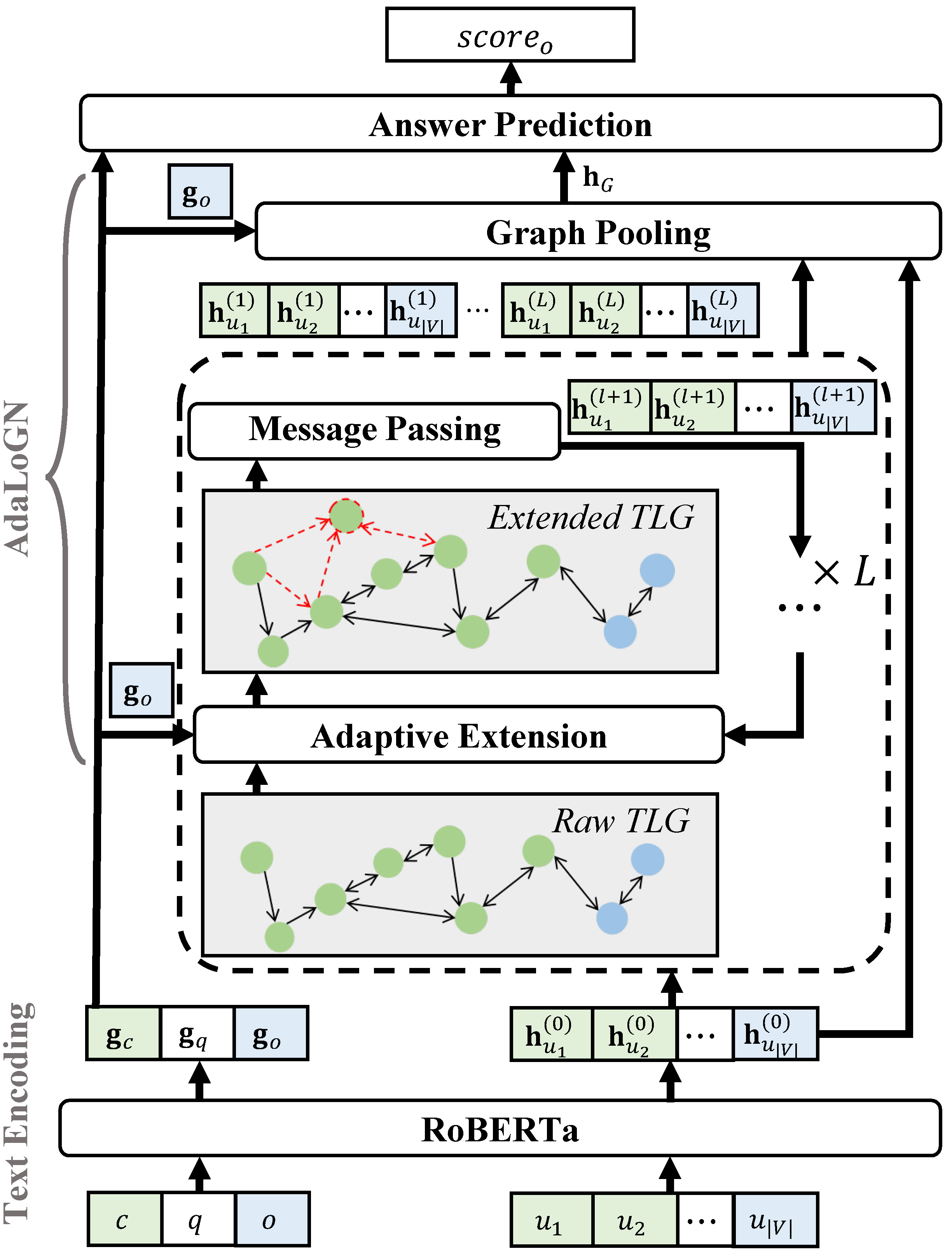}
    \caption{Overview of our approach.}
    \label{fig:model_architecture}
\end{figure}

Figure~\ref{fig:model_architecture} outlines our implementation. For each option $o \in O$, we generate the representations of~$c,q,o$ (i.e.,~$\mathbf{g}_c,\mathbf{g}_q,\mathbf{g}_o$, respectively) by a pre-trained language model (Section~\ref{sec:roberta}), and we construct a raw TLG where nodes (i.e.,~$u_1,\ldots,u_{|V|}$) represent EDUs extracted from~$c,q,o$ and edges represent their logical relations (Section~\ref{sec:tlg}). With their initial representations (i.e.,~$\mathbf{h}_{u_1}^{(0)},\ldots,\mathbf{h}_{u_{|V|}}^{(0)}$) obtained from the pre-trained language model, in an iterative manner, we adaptively extend the TLG (i.e.,~symbolic reasoning) and then pass messages (i.e.,~neural reasoning) to update node representations (i.e.,~$\mathbf{h}_{u_1}^{(l+1)},\ldots,\mathbf{h}_{u_{|V|}}^{(l+1)}$) for generating the representation of the TLG (i.e.,~$\mathbf{h}_G$) (Section~\ref{sec:adalogn}). Finally, we predict the correctness of~$o$ (i.e.,~$\mathit{score}_o$) based on the above representations (Section~\ref{sec:ans-pred}).

\subsection{Text Encoding}
\label{sec:roberta}

We use RoBERTa~\cite{roberta}, a pre-trained language model, to encode three token sequences $c = c_1 \cdots c_{|c|}$, $q = q_1 \cdots q_{|q|}$, and $o = o_1 \cdots o_{|o|}$ which are concatenated by the classifier token $<$s$>$ and the separator token $<$/s$>$:
\begin{equation}
\label{eq:enc-text-1}
\resizebox{\columnwidth}{!}{$
\begin{aligned}
    & [\mathbf{g}_\text{$<$s$>$}; \mathbf{g}_{c_1}; \ldots; \mathbf{g}_\text{$<$/s$>$}; \mathbf{g}_{q_1}; \ldots; \mathbf{g}_{o_1}; \ldots; \mathbf{g}_\text{$<$/s$>$}] \\
    = & ~\text{RoBERTa}(\text{$<$s$>$ $c_1$ $\cdots$ $<$/s$>$ $q_1$ $\cdots$ $o_1$ $\cdots$ $<$/s$>$}) \,.
\end{aligned}
$}
\end{equation}
\noindent The output vector representations are averaged to form the representations of~$c,q,o$:
\begin{equation}
\label{eq:enc-text-2}
\resizebox{\columnwidth}{!}{$
    \mathbf{g}_c = \frac{1}{|c|}\sum\limits_{i=1}^{|c|}{\mathbf{g}_{c_i}} \,,\;
    \mathbf{g}_q = \frac{1}{|q|}\sum\limits_{i=1}^{|q|}{\mathbf{g}_{q_i}} \,,\;
    \mathbf{g}_o = \frac{1}{|o|}\sum\limits_{i=1}^{|o|}{\mathbf{g}_{o_i}} \,.
$}
\end{equation}

\subsection{Text Logic Graph (TLG)}
\label{sec:tlg}

Besides directly encoding text, we extract logical relations from text as a graph called TLG.

\subsubsection{Definition of TLG}

For a piece of text, its TLG is a directed graph $G = \langle V, E \rangle$ where $V$~is a set of nodes representing EDUs of the text~\cite{RST}, and $E \subseteq V \times R \times V$ is a set of labeled directed edges representing \emph{logical relations} between EDUs described in the text. We consider six types of common logical relations $R = \{\texttt{conj}, \texttt{disj}, \texttt{impl}, \texttt{neg}, \texttt{rev}, \texttt{unk}\}$:
\begin{itemize}
    \item conjunction (\texttt{conj}), disjunction (\texttt{disj}), implication (\texttt{impl}), and negation (\texttt{neg}) are standard logical connectives in propositional logic;
    \item reversed implication (\texttt{rev}) is introduced to represent the inverse relation of \texttt{impl};
    \item \texttt{unk} represents an unknown relation.
\end{itemize}
\noindent Since \texttt{conj}, \texttt{disj}, \texttt{neg}, and \texttt{unk} are symmetric relations, edges labeled with them are bidirectional.

Observe the difference between our TLG and the discourse-based logic graph considered in DAGN~\cite{dagn}: edges in the former represent logical relations, while those in the latter represent discourse relations. Therefore, we can explicitly perform symbolic reasoning on TLG.

\begin{table}[t]
\small
    \centering
    \begin{tabular}{|p{4.5cm}|l|}
        \hline
        Rhetorical Relation & Logical Relation \\
        \hline
        LIST, CONTRAST & \texttt{conj} \\
        DISJUNCTION & \texttt{disj} \\
        RESULT & \texttt{impl} \\
        CAUSE, PURPOSE, CONDITION, BACKGROUND & \texttt{rev} \\
        \hline
    \end{tabular}
    \caption{Mapping from rhetorical relations in Graphene to logical relations in TLG.}
    \label{table:graphene_mapping}
\end{table}

\subsubsection{Construction of Raw TLG}
\label{sec:tlg-construction}

We initialize a raw TLG from~$c$ and~$o$. Following~\citet{dagn}, we ignore~$q$ as it is usually uninformative in existing datasets. Specifically, we use Graphene~\cite{DBLP:conf/coling/CettoNFH18} to extract EDUs and their rhetorical relations~\cite{RST} from~$c$ and~$o$. Rhetorical relations are converted to logical relations via the mapping in Table~\ref{table:graphene_mapping}. Note that each \texttt{impl} edge is always paired with an inverse \texttt{rev} edge, and vice versa.

We also define a small number of syntactic rules to identify EDUs that negate each other and connect them with \texttt{neg}. The rules are based on part-of-speech tags and dependencies. For example, one such rule checks whether two EDUs differ from each other only by an antonym of an adverb.

In addition, for each pair of EDUs that are adjacent in the text (including the last EDU of~$c$ and the first EDU of~$o$) but have none of the above logical relations, we connect them with \texttt{unk} because Graphene may fail to identify their relation.

\subsection{\MODELNAMEFULL\ (\MODELNAME)}
\label{sec:adalogn}

Since TLG consists of logical relations, we explicitly perform symbolic reasoning by applying inference rules to extend the TLG with inferred logical relations to benefit the subsequent neural reasoning. However, rather than computing the deductive closure which may undesirably provide many relations that are irrelevant to answering the question and mislead neural reasoning, we perform adaptive extension by leveraging signals from neural reasoning to identify and admit relevant extensions. For neural reasoning, we perform message passing to update node representations, which finally are pooled into the representation of the TLG to be used in the subsequent answer prediction. We iterate the above process by restarting inference on the raw TLG in each iteration with signals from the previous iteration to accommodate corrections to the reasoning process and let symbolic and neural reasoning sufficiently interact with each other. We transform the above idea into a new model named \MODELNAME\ outlined in Figure~\ref{fig:model_architecture} and detailed below.

\begin{figure}
     \centering
     \begin{subfigure}[h]{0.25\linewidth}
         \centering
         \includegraphics[width=0.9\textwidth]{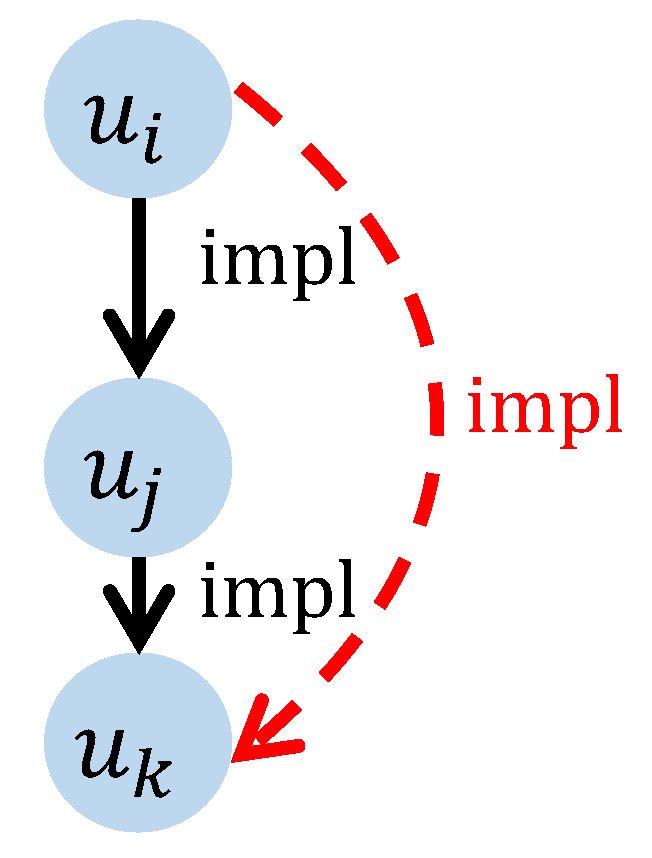}
         \caption{Hypothetical syllogism.}
         \label{fig:rule2}
     \end{subfigure}
     \hfill
     \begin{subfigure}[h]{0.3\linewidth}
         \centering
         \includegraphics[width=0.9\textwidth]{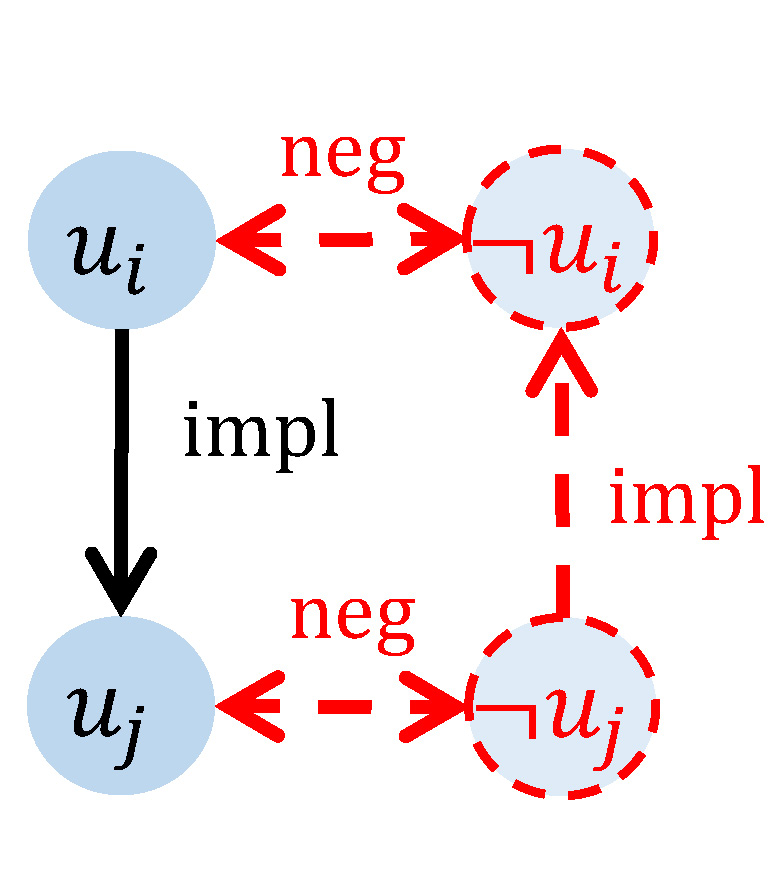}
         \caption{Transposition.}
         \label{fig:rule1}
     \end{subfigure}
     \hfill
     \begin{subfigure}[h]{0.25\linewidth}
         \centering
         \includegraphics[width=0.7\textwidth]{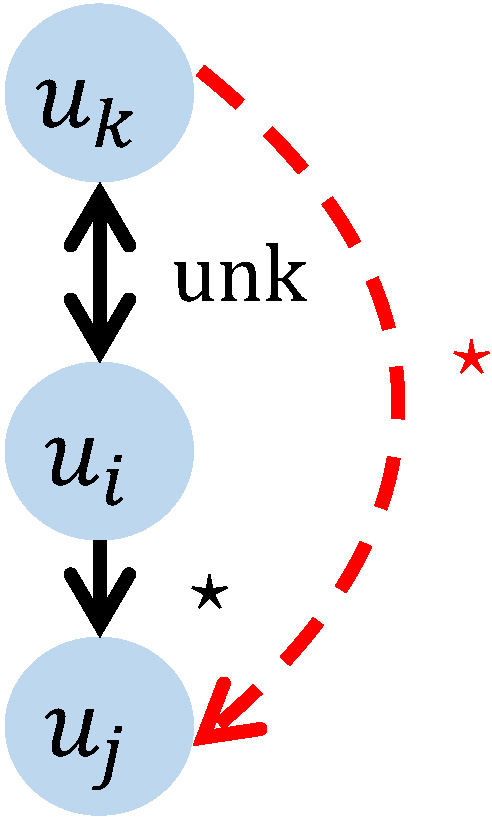}
         \caption{Adjacency-transmission.}
         \label{fig:rule3}
     \end{subfigure}
        \caption{Dashed nodes and edges are inferred by applying an inference rule. $\star$~represents any logical relation in $\{\texttt{conj},\texttt{disj},\texttt{impl}\}$. We omit \texttt{rev} edges.}
        \label{fig:substructure}
\end{figure}

\subsubsection{Inference Rules}

Let $G = \langle V, E \rangle$ be a raw TLG. For symbolic reasoning over the logical relations in~$G$, we apply two \emph{inference rules about implication in propositional logic}. Other rules are left for future work.
\begin{itemize}
    \item Hypothetical Syllogism:
    \begin{equation}
        ((u_i \rightarrow u_j) \wedge (u_j \rightarrow u_k)) \vdash (u_i \rightarrow u_k) \,.
    \end{equation}
    \noindent Specifically, if $E$~contains two edges $\langle u_i,\texttt{impl},u_j \rangle$ and $\langle u_j,\texttt{impl},u_k \rangle$, we can add two edges $\langle u_i,\texttt{impl},u_k \rangle$ and $\langle u_k,\texttt{rev},u_i \rangle$ to~$E$, as illustrated in Figure~\ref{fig:rule2}.
    \item Transposition:
    \begin{equation}
        (u_i \rightarrow u_j) \vdash (\neg u_j \rightarrow \neg u_i) \,.
    \end{equation}
    \noindent Specifically, if $E$~contains an edge $\langle u_i,\texttt{impl},u_j \rangle$, we can add two edges $\langle \neg u_j, \texttt{impl}, \neg u_i \rangle$ and $\langle \neg u_i, \texttt{rev}, \neg u_j \rangle$ to~$E$, as illustrated in Figure~\ref{fig:rule1}. Note that if~$u_i$ (resp.~$u_j$) is not incident from/to any $\texttt{neg}$ edge, i.e.,~$\neg u_i$ (resp. $\neg u_j$) is not a node in~$V$, we will add~$\neg u_i$ (resp. $\neg u_j$) to~$V$ whose text negates that of~$u_i$ (resp.~$u_j$),
    and then add a bidirectional \texttt{neg} edge between~$u_i$ and $\neg u_i$ (resp.~$u_j$ and $\neg u_j$) to~$E$.
\end{itemize}

Besides, recall that \texttt{unk} represents a potential logical relation between EDUs that are adjacent in text. Considering that an EDU often inherits logical relations from its adjacent EDUs, we heuristically define and apply the following inference rule.
\begin{itemize}
    \item Adjacency-Transmission:
    \begin{equation}
        ((u_i \star u_j) \wedge (u_i \sim u_k)) \vdash (u_k \star u_j) \,,
    \end{equation}
    \noindent where $\star \in \{\wedge,\vee,\rightarrow\}$ and $\sim$~represents adjacency in text. For example, if $E$~contains two edges $\langle u_i,\texttt{conj},u_j \rangle$ and $\langle u_i,\texttt{unk},u_k \rangle$, we can add a bidirectional \texttt{conj} edge between~$u_k$ and $u_j$ to~$E$, as illustrated in Figure~\ref{fig:rule3}.
\end{itemize}
\noindent While this rule may generate false propositions, we expect our adaptive reasoner to apply it properly. For example, it is useful for handling the following sentence: ``... only 1~person in the group knew 3~of the group~($u_k$), 3~people knew 2~of the group~($u_i$), and 4~people know 1~of the group~($u_j$).'' Graphene identifies $\langle u_i,\texttt{conj},u_j \rangle$ and $\langle u_i,\texttt{unk},u_k \rangle$ but misses $\langle u_k,\texttt{conj},u_j \rangle$, which can be generated by applying this rule.

\subsubsection{Adaptive Extension of TLG}

Our symbolic reasoning is \emph{adaptive}. We rely on signals from neural reasoning to decide which inference steps are relevant to answering the questions and hence are admitted to extend the TLG. Specifically, each candidate extension~$\epsilon$ applies an inference rule over a set of nodes $V_\epsilon \subseteq V$. We average their vector representations (which will be detailed later) to form the representation of~$\epsilon$:
\begin{equation}
\label{eq:enc-ext}
    \mathbf{h}_{\epsilon} = \frac{1}{|V_\epsilon|}\sum_{u_i \in V_\epsilon}{\mathbf{h}_{u_i}} \,.
\end{equation}
\noindent Since $\epsilon$~is for predicting the correctness of~$o$, we interact~$\mathbf{h}_{\epsilon}$ with the representation of~$o$, i.e.,~$\mathbf{g}_o$ in Equation~(\ref{eq:enc-text-2}), to predict the relevance score of~$\epsilon$:
\begin{equation}
\label{eq:rel}
    \mathit{rel}_\epsilon = \sigmoid(\linear(\mathbf{h}_{\epsilon} \parallel \mathbf{g}_o)) \,,
\end{equation}
\noindent where $\parallel$~represents vector concatenation.
We admit all possible~$\epsilon$ to extend~$G$ such that $\mathit{rel}_\epsilon > \tau$ where $\tau$~is a predefined threshold.

Moreover, our neural-symbolic reasoning is \emph{iterative}. In the $(l+1)$-th iteration, we restart symbolic reasoning with the raw TLG and recompute Equation~(\ref{eq:enc-ext}) with node representations~$\mathbf{h}_{u_i}^{(l)}$ from neural reasoning in the $l$-th iteration (which will be detailed in Section~\ref{sec:adalogn-mp}). The initial node representations~$\mathbf{h}_{u_i}^{(0)}$ are obtained from a pre-trained language model. Specifically, we flatten~$V$ into a sequence of nodes in the order they appear in the text. Recall that $V$~is divided into $V_c=\{u_1,\ldots,u_{|V_c|}\}$ and $V_o=\{u_{|V_c|+1},\ldots,u_{|V|}\}$ representing the nodes extracted from~$c$ and~$o$, respectively. Each node~$u_i$ is a token sequence $u_i = u_{i_1} \cdots u_{i_{|u_i|}}$. We use RoBERTa to encode~$V_c$ and~$V_o$ which are concatenated by $<$s$>$ and $<$/s$>$, where nodes inside~$V_c$ and~$V_o$ are separated by a special token ``$\mid$'':
\begin{equation}
\label{eq:enc-raw-1}
\resizebox{\columnwidth}{!}{$
\begin{aligned}
    & [\mathbf{h}_\text{$<$s$>$}; \mathbf{h}_{u_{1_1}}; \ldots; \mathbf{h}_\mid; \ldots; \mathbf{h}_\text{$<$/s$>$}; \mathbf{h}_{u_{|V_c|+1_1}}; \ldots; \mathbf{h}_\mid; \ldots; \mathbf{h}_\text{$<$/s$>$}] \\
    = & ~\text{RoBERTa}(\text{$<$s$>$ $u_{1_1} \cdots \mid \cdots$ $<$/s$>$ $u_{|V_c|+1_1} \cdots \mid \cdots$ $<$/s$>$}) \,.
\end{aligned}
$}
\end{equation}
\noindent The output vector representations are averaged to form the initial representation of each node $u_i \in V$:
\begin{equation}
\label{eq:enc-raw-2}
    \mathbf{h}_{u_i}^{(0)} = \frac{1}{|u_i|}\sum_{j=1}^{|u_i|}{\mathbf{h}_{u_{i_j}}} \,.
\end{equation}

\subsubsection{Message Passing}
\label{sec:adalogn-mp}

To let the nodes in TLG interact with each other and fuse their information, our neural reasoning performs graph-based message passing~\cite{mpnn} to update node representations in each iteration from~$\mathbf{h}_{u_i}^{(l)}$ to~$\mathbf{h}_{u_i}^{(l+1)}$. Since TLG is a heterogeneous graph containing multiple types of edges, we incorporate the node-to-node message passing mechanism in R-GCN~\cite{rgcn} as a basis. Further, observe that TLG is usually loosely connected and prone to cause insufficient interaction between~$V_c$ and~$V_o$ via long paths in limited iterations, which cannot be alleviated by simply increasing the number of iterations because it would raise other issues such as over-smoothing~\cite{DBLP:conf/aaai/LiHW18,DBLP:conf/aaai/ChenLLLZS20}. To enhance such interaction which is critical to predicting the correctness of~$o$, we incorporate a novel \emph{subgraph-to-node message passing mechanism} to holistically pass the information aggregated from a subgraph (e.g.,~$V_c$) to a node (e.g.,~each $u_i \in V_o$).

Specifically, without loss of generality, for each $u_i \in V_o$, we compute the $u_i$-attended aggregate representation of~$V_c$ by an attention-weighted sum of node representations over~$V_c$:
\begin{align}
\begin{split}
    \mathbf{h}_{V_c,u_i}^{(l)} & = \sum_{u_j \in V_c}{\alpha_{i,j} \mathbf{h}_{u_j}^{(l)}} \,, \text{ where} \\
    \alpha_{i,j} & = \softmax_j([a_{i,1};\ldots;a_{i,|V_c|}]^\intercal) \,, \\
    a_{i,j} & = \leakyrelu(\linear(\mathbf{h}_{u_i}^{(l)} \parallel \mathbf{h}_{u_j}^{(l)})) \,.
\end{split}
\end{align}
\noindent Let~$N^i$ be the set of neighbors of~$u_i$. Let $N_r^i \subseteq N^i$ be the subset under logical relation $r \in R$. We update the representation of~$u_i$ by passing messages to~$u_i$ from its neighbors and from~$V_c$:
\begin{equation}
\label{eq:mp}
\resizebox{\columnwidth}{!}{$
\begin{aligned}
    \mathbf{h}_{u_i}^{(l+1)} & = \relu(
    \sum\limits_{r \in R}{\sum\limits_{u_j \in N_r^i}{\frac{\alpha_{i,j}}{|N_r^i|} \mathbf{W}_r^{(l)} \mathbf{h}_{u_j}^{(l)} }}
    + \mathbf{W}_0^{(l)} \mathbf{h}_{u_i}^{(l)} \\
    & \qquad\qquad + \beta_i \mathbf{W}_\text{subgraph}^{(l)} \mathbf{h}_{V_c,u_i}^{(l)} ) \,, \text{ where} \\
    \alpha_{i,j} & = \softmax_{\idx(a_{i,j})}([\ldots;a_{i,j};\ldots]^\intercal) \text{ for all } u_j \in N^i \,, \\
    a_{i,j} & = \leakyrelu(\linear(\mathbf{h}_{u_i}^{(l)} \parallel \mathbf{h}_{u_j}^{(l)})) \,, \\
    \beta_i & = \sigmoid(\linear(\mathbf{h}_{u_i}^{(l)} \parallel \mathbf{h}_{V_c,u_i}^{(l)})) \,,
\end{aligned}
$}
\end{equation}
\noindent $\mathbf{W}_r^{(l)}, \mathbf{W}_0^{(l)}, \mathbf{W}_\text{subgraph}^{(l)}$ are matrices of learnable parameters, and $\idx(a_{i,j})$ returns the index of~$a_{i,j}$ in the $|N^i|$-dimensional vector $[\ldots;a_{i,j};\ldots]^\intercal$.

In an analogous way, for each $u_i \in V_c$, we compute the $u_i$-attended aggregate representation of~$V_o$ denoted by~$\mathbf{h}_{V_o,u_i}^{(l)}$ and update $\mathbf{h}_{u_i}^{(l+1)}$.

Observe two differences between Equation~(\ref{eq:mp}) and its counterpart in the original R-GCN. First, we incorporate subgraph-to-node message passing and control it by a gating mechanism (i.e.,~$\beta_i$). Second, we weight node-to-node message passing by an attention mechanism (i.e.,~$\alpha_{i,j}$).

\subsubsection{Graph Pooling}

After $L$~iterations where $L$~is a hyperparameter, for each node $u_i \in V$, we fuse its representations over all the iterations with a residual connection:
\begin{equation}
    \mathbf{h}_{u_i}^\text{fus} = \mathbf{h}_{u_i}^{(0)} + \linear(\mathbf{h}_{u_i}^{(1)} \parallel \cdots \parallel \mathbf{h}_{u_i}^{(L)}) \,.
\end{equation}
\noindent Inspired by~\citet{dagn}, we feed all~$\mathbf{h}_{u_i}^\text{fus}$ into a bidirectional residual GRU layer~\cite{gru} to finalize node representations:
\begin{equation}
    [\mathbf{h}_{u_1}^\text{fnl}; \ldots; \mathbf{h}_{u_{|V|}}^\text{fnl}] = \resbigru([\mathbf{h}_{u_1}^\text{fus}; \ldots; \mathbf{h}_{u_{|V|}}^\text{fus}]) \,.
\end{equation}
We aggregate these node representations by computing an $o$-attended weighted sum:
\begin{align}
\begin{split}
    \mathbf{h}_V & = \sum_{u_i \in V}{\alpha_i \mathbf{h}_{u_i}^\text{fnl}} \,, \text{ where} \\
    \alpha_i & = \softmax_i([a_1;\ldots;a_{|V|}]^\intercal) \,, \\
    a_i & = \leakyrelu(\linear(\mathbf{g}_o \parallel \mathbf{h}_{u_i}^\text{fnl})) \,,
\end{split}
\end{align}
\noindent and~$\mathbf{g}_o$ is the representation of~$o$ in Equation~(\ref{eq:enc-text-2}). We concatenate~$\mathbf{h}_V$ and the relevance scores to form the representation of~$G$:
\begin{align}
\begin{split}
    \mathbf{h}_G & = (\mathbf{h}_V \parallel \mathit{rel}_{\mathcal{E}^{(1)}} \parallel \cdots \parallel \mathit{rel}_{\mathcal{E}^{(L)}}) \,, \text{ where} \\
    \mathit{rel}_{\mathcal{E}^{(l)}} & = \frac{1}{|\mathcal{E}^{(l)}|} \sum_{\epsilon \in \mathcal{E}^{(l)}}{\mathit{rel}_\epsilon} \,,
\end{split}
\end{align}
\noindent $\mathcal{E}^{(l)}$~is the set of candidate extensions in the $l$-th iteration, and $\mathit{rel}_\epsilon$ is in Equation~(\ref{eq:rel}). In this way, we are able to train the network in Equation~(\ref{eq:rel}).

\subsection{Answer Prediction}
\label{sec:ans-pred}

We fuse the representations of $c,q,o$ and the TLG to predict the correctness of~$o$:
\begin{equation}
\resizebox{\columnwidth}{!}{$
    \mathit{score}_o = \linear(\tanh(\linear(\mathbf{g}_c \parallel \mathbf{g}_q \parallel \mathbf{g}_o \parallel \mathbf{h}_G))) \,,
$}
\end{equation}
\noindent where $\mathbf{g}_c, \mathbf{g}_q, \mathbf{g}_o$ are in Equation~(\ref{eq:enc-text-2}).

\subsection{Loss Function}

Let $o_\text{gold} \in O$ be the correct answer. We optimize the cross-entropy loss with label smoothing:
\begin{align}
\label{eq:loss}
\begin{split}
    \mathcal{L} & = -(1-\gamma)\mathit{score}'_{o_\text{gold}} - \gamma \frac{1}{|O|}\sum_{o_i \in O}{\mathit{score}'_{o_i}} \,, \\ \text{where}~ &
    \mathit{score}'_{o_i} = \log\frac{\exp(\mathit{score}_{o_i})}{\sum_{o_j \in O}{\exp(\mathit{score}_{o_j})}} \,,
\end{split}
\end{align}
\noindent and $\gamma$~is a predefined smoothing factor.

%% file: 4-experiments.tex
\section{Experiments}
\label{sec:experiments}

\subsection{Datasets}

We used two reasoning-based MRC datasets.

ReClor~\cite{reclor} consists of 6,138~four-option multiple-choice questions collected from standardized exams such as GMAT and LSAT. The questions were divided into 4,638~for training, 500~for development, and 1,000~for testing. The test set was further divided into 440~easy questions (Test-E) where each question could be correctly answered by some strong baseline method using only the options and ignoring the context and the question, and the rest 560~hard questions (Test-H).

LogiQA~\cite{logiqa} consists of 8,768~four-option multiple-choice questions collected from the National Civil Servants Examination of China, which were translated into English. The questions were divided into 7,376~for training, 651~for development, and 651~for testing.

\subsection{Implementation Details}

We experimented on NVIDIA V100 (32GB).

We tuned hyperparameters on the development set of each dataset. Specifically, for text encoding, we used RoBERTa-large with $\text{hidden layer}=24$ and $\text{hidden units}=1,024$ implemented by Hugging Face~\cite{huggingface}. For message passing, our implementation was based on DGL~\cite{dgl}. For both datasets, we used the Adam optimizer, and set $\text{attention heads}=16$, $\text{dropout rate}=0.1$, $\text{epochs}=10$, $\text{batch size}=16$ selected from $\{8,16,24\}$, number of iterations $L=2$ from $\{2,3\}$, and $\text{maximum sequence length}=384$.
For ReClor, we set $\text{warm-up proportion}=0.1$ from $\{0.1,0.2\}$, $\text{learning rate}=7e\text{--}6$ from $\{6e\text{--}6, 7e\text{--}6, 8e\text{--}6, 1e\text{--}5\}$, and $\text{seed}=123$ from $\{123, 1234, 42, 43\}$. For LogiQA, we set $\text{warm-up proportion}=0.2$ from $\{0.1,0.2\}$, $\text{learning rate}=8e\text{--}6$ from $\{6e\text{--}6, 7e\text{--}6, 8e\text{--}6, 1e\text{--}5\}$, and $\text{seed}=42$ from $\{123, 1234, 42, 43\}$.

For the relevance score threshold~$\tau$ below Equation~(\ref{eq:rel}), we set $\tau=0.6$ from $\{0.4, 0.5, 0.6, 0.7\}$ for both datasets. For the smoothing factor~$\gamma$ in Equation~(\ref{eq:loss}), we set $\gamma=0.25$ for both datasets.

To fit in our GPU's memory, we restricted a raw TLG to contain at most 25~nodes and 50~edges by, if needed, randomly merging nodes connected by an \texttt{unk} edge and/or deleting non-bridge edges while keeping the graph connected.

\subsection{Baselines}

We compared our approach, referred to as \MODELNAME, with popular pre-trained language models and with other known methods in the literature.

Reasoning-based MRC, like other MRC tasks, can be solved by using a pre-trained language model with a classification layer. \citet{reclor}~reported the results of BERT$_\texttt{LARGE}$, RoBERTa$_\texttt{LARGE}$, and XLNet$_\texttt{LARGE}$ on ReClor. \citet{dagn}~reported the results of BERT$_\texttt{LARGE}$ and RoBERTa$_\texttt{LARGE}$ on LogiQA.

In the literature, we found the results of DAGN~\cite{dagn}, Focal Reasoner~\cite{facolreasoner}, and LReasoner~\cite{lreasoner2,lreasoner1} on both datasets. For a fair comparison with our approach, we presented their results on RoBERTa$_\texttt{LARGE}$, while LReasoner achieved better results with ALBERT. Between the two variants of LReasoner, one without data augmentation (w/o~DA) and the other with data augmentation (w/~DA), we presented both of their results but mainly compared with the former because our approach and other baseline methods would also benefit if data augmentation were incorporated.

\subsection{Evaluation Metric}

Following the literature, we reported accuracy, i.e.,~the proportion of correctly answered questions. For our approach we reported the max across 3~runs on the development set of each dataset.

\begin{table}[t]
    \centering
    \small
    \begin{tabular}{|l|cccc|}
        \hline
        Method & Dev & Test & Test-E & Test-H  \\
        \hline
        BERT$_\texttt{LARGE}$ & 53.80 & 49.80 & 72.00 & 32.30\\
        RoBERTa$_\texttt{LARGE}$ & 62.60 & 55.60 & 75.50 & 40.00  \\
        XLNet$_\texttt{LARGE}$ & 62.00 & 56.00 & 75.70 & 40.50 \\
        DAGN & 65.80 & 58.30 & 75.91 & 44.46 \\
        Focal Reasoner & 66.80  & 58.90 & 77.05 & 44.64 \\
        LReasoner~(w/o~DA) & 65.20 & 58.30 & 78.60 & 42.30 \\
        LReasoner~(w/~DA) & 66.20 & 62.40 & 81.40 & 47.50 \\
        \MODELNAME & 65.20 & 60.20 & 79.32 & 45.18 \\
        \hline
        Human & -- & 63.00 & 57.10 & 67.20 \\
        \hline
    \end{tabular}
    \caption{Comparison with baselines on ReClor.}
    \label{table:main_results_reclor}
\end{table}

\begin{table}[t]
    \centering
    \small
    \begin{tabular}{|l|cc|}
        \hline
        Method & Dev & Test \\
        \hline
        BERT$_\texttt{LARGE}$ & 34.10 & 31.03 \\
        RoBERTa$_\texttt{LARGE}$ & 35.02 & 35.33 \\
        DAGN & 36.87 & 39.32 \\
        Focal Reasoner & 41.01 & 40.25\\
        LReasoner~(w/~DA) & 38.10 & 40.60 \\
        \MODELNAME  & 39.94  & 40.71  \\
        \hline
        Human & -- & 86.00 \\
        \hline
    \end{tabular}
    \caption{Comparison with baselines on LogiQA.}
    \label{table:main_results_logiqa}
\end{table}

\subsection{Comparison with Baselines}

On ReClor, as shown in Table~\ref{table:main_results_reclor}, \MODELNAME\ outperformed all the baseline methods on the test set by at least~1.30\%, except for LReasoner~(w/~DA) which performed data augmentation so that the comparison might be unfair. \MODELNAME\ and LReasoner~(w/~DA) both exceeded~60\%, being comparable with human-level performance~(63\%).

On LogiQA, as shown in Table~\ref{table:main_results_logiqa}, \MODELNAME\ outperformed all the baseline methods on the test set, including LReasoner~(w/~DA). Still, our result~(40.71\%) was not comparable with human-level performance~(86\%).

In particular, on both ReClor and LogiQA, \MODELNAME\ exceeded DAGN on the test set by 1.39\%--1.90\%, which demonstrated the effectiveness of our approach in addressing the limitations of DAGN mentioned in Section~\ref{sec:intro}.

\begin{table}[t]
    \centering
    \small
    \begin{tabular}{|l|cccc|}
        \hline
        Method & Dev & Test & Test-E & Test-H  \\
        \hline
        \MODELNAME & 65.20 & 60.20 & 79.32 & 45.18 \\
        \MODELNAME$_\texttt{no-ext}$ & 65.80  & 59.50 &  77.27 & 45.54 \\
        \MODELNAME$_\texttt{full-ext}$ & 65.00  & 58.80 & 78.19  & 43.57 \\
        \MODELNAME$_\texttt{no-at}$ & 64.80 & 59.40 & 79.77 & 43.39 \\
        \MODELNAME$_\texttt{n2n}$ & 65.20 & 57.60 & 77.95 & 41.61 \\
        \MODELNAME$_\texttt{n2n+}$ & 65.00  & 58.60  & 78.64 &  42.86 \\
        \hline
    \end{tabular}
    \caption{Ablation study on ReClor.}
    \label{table:ablation_results_reclor}
\end{table}

\begin{table}[t]
    \centering
    \small
    \begin{tabular}{|l|cc|}
        \hline 
        Method & Dev & Test\\
        \hline
        \MODELNAME & 39.94  & 40.71 \\
        \MODELNAME$_\texttt{no-ext}$ & 37.94 & 39.02 \\
        \MODELNAME$_\texttt{full-ext}$ & 39.63 & 39.02 \\
        \MODELNAME$_\texttt{no-at}$ & 38.56 & 39.94 \\
        \MODELNAME$_\texttt{n2n}$ & 38.40 & 39.02 \\
        \MODELNAME$_\texttt{n2n+}$ & 38.40 & 38.86 \\
        \hline
    \end{tabular}
    \caption{Ablation study on LogiQA.}
    \label{table:ablation_results_logiqa}
\end{table}

\subsection{Ablation Study}

We conducted an ablation study to evaluate the effectiveness of the two main technical contributions in our approach: adaptive extension of TLG and subgraph-to-node message passing.

\subsubsection{Effectiveness of Adaptive Extension}

We compared the standard version of \MODELNAME\ with two variants removing adaptive extension.
\begin{itemize}
    \item \MODELNAME$_\texttt{no-ext}$ performs no extension.
    \item \MODELNAME$_\texttt{full-ext}$ performs full extension by computing and admitting the deductive closure.
\end{itemize}

On ReClor, as shown in Table~\ref{table:ablation_results_reclor}, both variants exhibited a fair decrease in accuracy on the test set by 0.70\%--1.40\%. On LogiQA, as shown in Table~\ref{table:ablation_results_logiqa}, the decreases were larger, 1.69\% on the test set, possibly because the questions in LogiQA were harder so that the effectiveness of our adaptive extension became more noticeable. Interestingly, on both datasets, \MODELNAME$_\texttt{full-ext}$ was not better than \MODELNAME$_\texttt{no-ext}$ on the test set, indicating that a naive injection of logical reasoning into neural reasoning might not have positive effects.

\begin{figure}[t]
    \centering
    \includegraphics[width=0.9\linewidth]{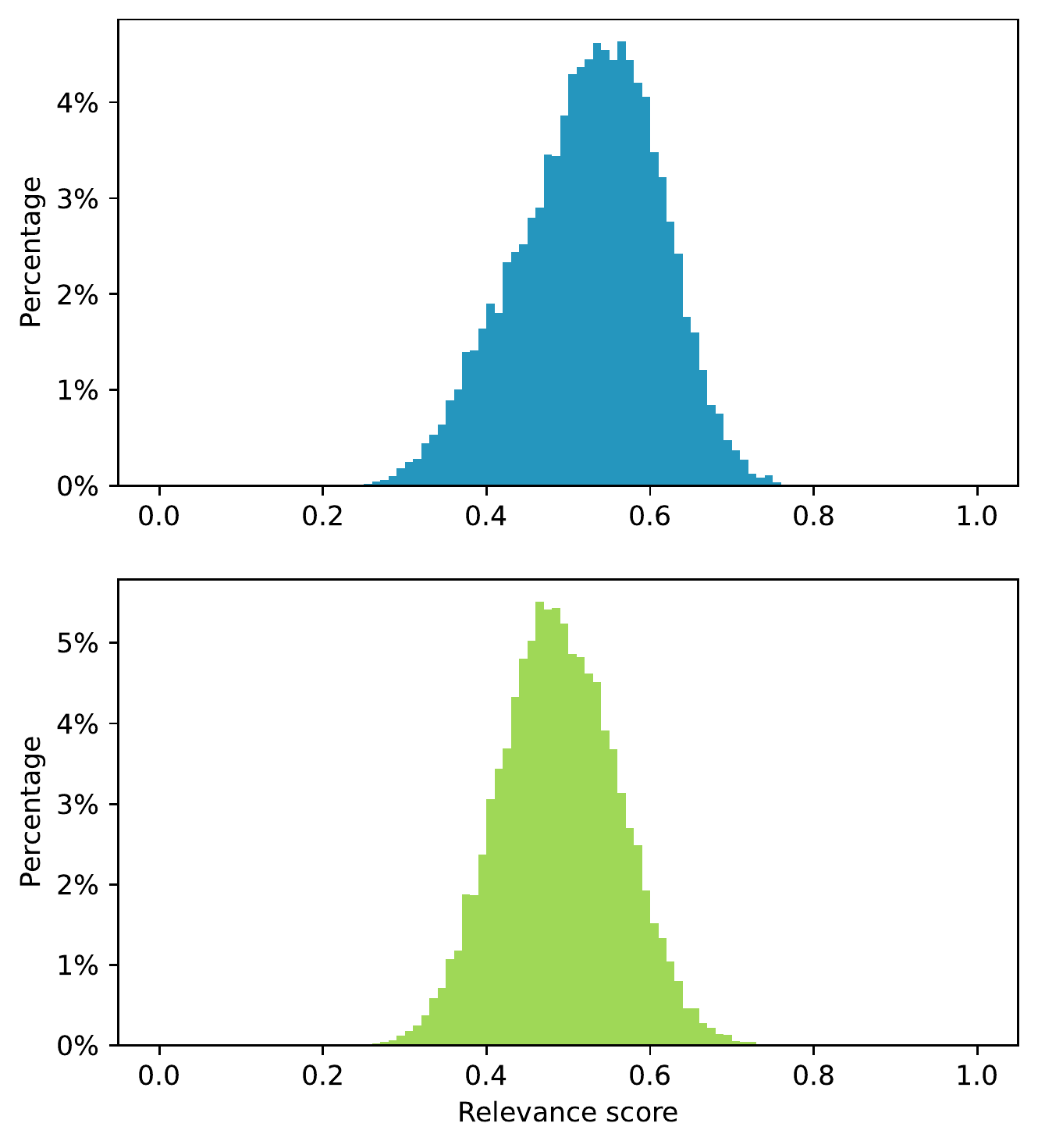}
    \caption{Distributions of relevance scores of candidate extensions. Top: on the development set of Reclor; Bottom: on the development set of LogiQA.}
    \label{fig:rel_score_dist}
\end{figure}

We analyzed the distributions of relevance scores of candidate extensions, i.e.,~$\mathit{rel}_\epsilon$ in Equation~(\ref{eq:rel}). As shown in Figure~\ref{fig:rel_score_dist}, they approximated a normal distribution on both datasets. By setting the threshold $\tau=0.6$, we admitted 19.57\% and~4.86\% of the extensions on ReClor and LogiQA, respectively.

We also compared with a variant of \MODELNAME\ using a subset of inference rules.
\begin{itemize}
    \item \MODELNAME$_\texttt{no-at}$ ignores the adjacency-transmission rule.
\end{itemize}

By ignoring the adjacency-transmission rule, \MODELNAME$_\texttt{no-at}$ showed a decrease in accuracy on the test sets by 0.77\%--0.80\%, suggesting the usefulness of this rule despite its heuristic nature.

\subsubsection{Effectiveness of Subgraph-to-Node Message Passing}

We compared the standard version of \MODELNAME\ with two variants removing subgraph-to-node message passing or implementing it in a different way.
\begin{itemize}
    \item \MODELNAME$_\texttt{n2n}$ only performs node-to-node message passing in a standard way.
    \item \MODELNAME$_\texttt{n2n+}$ only performs node-to-node message passing but, as an alternative to our holistic subgraph-to-node message passing, it adds a bidirectional \texttt{unk} edge between each node in the context subgraph and each node in the option subgraph to enhance context-option interaction.
\end{itemize}

On ReClor, as shown in Table~\ref{table:ablation_results_reclor}, both variants exhibited a large decrease in accuracy on the test set by 1.60\%--2.60\%. On LogiQA, as shown in Table~\ref{table:ablation_results_logiqa}, the decreases were also large, 1.69\%--1.85\% on the test set. The results demonstrated the effectiveness of our subgraph-to-node message passing.

Compared with \MODELNAME$_\texttt{n2n}$, \MODELNAME$_\texttt{n2n+}$ achieved better results on ReClor but worse results on LogiQA on the test set, indicating that a naive enhancement of context-option interaction could have negative effects.

\subsection{Error Analysis}

From the development set of each dataset, we randomly sampled fifty questions to which our approach outputted an incorrect answer. We analyzed the sources of these errors. Note that an error could have a mixture of multiple sources.

As shown in Table~\ref{table:error_analysis}, we mainly relied on Graphene to extract a raw TLG from text based on syntactic analysis, which accounted for about one third of the errors (36\%--38\%). Our adaptive extension of TLG constituted about one fifth of the errors (18\%--22\%), e.g.,~some excessive extensions produced irrelevant logical relations which might mislead message passing. One fifth of the errors (18\%--20\%) were due to the limited expressivity of our symbolic reasoning, i.e.,~a subset of propositional logic, while some questions required quantifiers. Other errors might be related to neural reasoning such as message passing or answer prediction (40\%--46\%).

\begin{table}[t]
    \centering
    \small
    \begin{tabular}{|l|r|r|}
        \hline
        Source of Error & ReClor & LogiQA \\
        \hline
         Construction of raw TLG & 38\% & 36\% \\
         Adaptive extension of TLG & 18\% & 22\% \\
         Expressivity of symbolic reasoning & 20\% & 18\% \\
         Others (about neural reasoning) & 46\% & 40\% \\
        \hline
    \end{tabular}
    \caption{Error analysis of \MODELNAME.}
    \label{table:error_analysis} 
\end{table}

\subsection{Run Time}

On both ReClor and LogiQA, our approach used about 0.8~second for answering a question.

%% file: 4.5-related_work.tex
\section{Related Work}
\label{sec:related_work}

\subsection{Reasoning-Based MRC}

While simple MRC tasks have been well studied, complex MRC tasks requiring various reasoning capabilities are receiving increasing research attention.
Among others, multi-hop MRC tasks in HotpotQA~\cite{hotpotqa} and WikiHop~\cite{wikihop} require retrieving and reading multiple supporting passages to answer a question. They can be solved by constructing and reasoning over a graph connecting passages that overlap or co-occur with each other~\cite{DBLP:conf/acl/QiuXQZLZY19,DBLP:conf/aaai/TuHW0HZ20}, by implicitly supervising a retriever via word weighting~\cite{DBLP:conf/emnlp/HuangWS0Q21}, or by iteratively applying dense retrieval~\cite{DBLP:conf/iclr/XiongLIDLWMY0KO21}.
MRC tasks in DROP~\cite{dropdataset} require discrete reasoning such as addition, counting, and sorting. Neural networks have been extended to incorporate modules that can perform such reasoning over numbers and dates mentioned in a given context~\cite{DBLP:conf/iclr/GuptaLR0020}.
For MRC tasks in CommonsenseQA~\cite{commonsenseqa} which are targeted at commonsense knowledge and reasoning, recent methods fuse external commonsense knowledge with pre-trained language models for reasoning~\cite{DBLP:conf/acl/YanRCZRZKLR21,DBLP:conf/acl/XuZXLZH21}. There are also studies on MRC tasks requiring spatial/geographical reasoning \cite{DBLP:conf/emnlp/HuangSLWCZDQ19,DBLP:conf/aaai/LiS021} and temporal/causal reasoning~\cite{DBLP:conf/coling/SunCQ18}.

Different from the above reasoning capabilities, the MRC tasks considered in this paper require \emph{logical reasoning}, such as reasoning about sufficient and necessary conditions, categorization, conjunctions and disjunctions. Pre-trained language models alone struggled and were far behind human-level performance on such tasks in ReClor~\cite{reclor} and LogiQA~\cite{logiqa} due to their weakness in logical reasoning.

Among existing methods for solving such tasks, DAGN~\cite{dagn} and Focal Reasoner~\cite{facolreasoner} extract discourse or coreference relations from text and represent as a graph of text units. Then they employ GNN to pass messages and update representations for predicting an answer. Different from their neural nature, our approach \emph{symbolically performs logical reasoning} as required by such tasks, by applying inference rules over extracted logical relations to extend the graph.
This feature resembles LReasoner~\cite{lreasoner2,lreasoner1} which extends the context with inferred logical relations to benefit the subsequent neural reasoning.
However, different from LReasoner which computes the deductive closure and identifies relevant extensions by text overlapping with the options in an unsupervised manner, our approach predicts relevance based on signals from neural reasoning \emph{in a supervised manner}, and our prediction \emph{evolves over iterations} after sufficient interaction between symbolic and neural reasoning. All these features helped our approach achieve better performance in the experiments.

\subsection{Neural-Symbolic Reasoning}

Our approach represents a novel implementation of neural-symbolic reasoning~\cite{DBLP:conf/ijcai/RaedtDMM20}, and it differs from the following existing methods.

One paradigm of neural-symbolic reasoning is logic-driven neural reasoning. For example, logical constraints can be compiled into a neural network by augmenting the loss function~\cite{DBLP:conf/icml/XuZFLB18} or the network structure~\cite{DBLP:conf/acl/LiS19}. Logical connectives, quantifiers, and consistency checking can also be approximated by neural networks~\cite{DBLP:conf/iclr/DongMLWLZ19,DBLP:conf/iclr/RenHL20,GGNN}. While these methods incorporate logical reasoning into neural reasoning via emulation, our approach \emph{explicitly performs logical reasoning} by applying inference rules over logical relations. Such exact inference is more accurate than emulation-based approximation.

Another paradigm is neural-driven logical reasoning. For example, neural networks have been employed to predict the truth of an atom in answering first-order logic queries~\cite{DBLP:conf/iclr/ArakelyanDMC21}, and to implement predicates in probabilistic logic programming~\cite{DBLP:journals/ai/ManhaeveDKDR21}. These methods and our approach cope with different problems, thus using different techniques. Specifically, while these methods complement logical reasoning with extra facts generated by neural reasoning, our approach \emph{filters inferred logical relations} based on signals from neural reasoning.

Moreover, observe that the neural-symbolic interaction in the above methods are unidirectional, i.e.,~they leverage either symbolic or neural reasoning to reinforce the other. By contrast, we allow \emph{bidirectional neural-symbolic interaction} where neural and symbolic reasoning mutually and iteratively reinforce each other for better performance.

%% file: 5-conclusion.tex
\section{Conclusion}
\label{sec:conclusion}

To meet the challenge of reasoning-based MRC, we presented a neural-symbolic approach where neural and symbolic reasoning mutually and iteratively reinforce each other via our new \MODELNAME\ model. We also enhanced graph-based neural reasoning with a novel subgraph-to-node message passing mechanism. Since these ideas are quite general, we believe they have great potential for a variety of applications beyond MRC, e.g.,~link prediction.

Error analysis has revealed some shortcomings of our approach. Currently we rely on syntactic tools to extract a raw TLG from text. We will explore other extraction methods to achieve a higher quality. We also plan to apply more inference rules and incorporate quantifiers to improve the expressivity of our symbolic reasoning.